# Automatic Identification of Types of Alterations in Historical Manuscripts

David Lassner[a], Anne Baillot[b], Sergej Dogadov[c], Klaus-Robert Müller[d], Shinichi Nakajima[e]

## Abstract

Alterations in historical manuscripts such as letters represent a promising field of research. On the one hand, they help understand the construction of text. On the other hand, topics that are being considered sensitive at the time of the manuscript gain coherence and contextuality when taking alterations into account, especially in the case of deletions. The analysis of alterations in manuscripts, though, is a traditionally very tedious work. In this paper, we present a machine learning-based approach to help categorize alterations in documents. In particular, we present a new probabilistic model (Alteration Latent Dirichlet Allocation, *alterLDA* in the following) that categorizes content-related alterations. The method proposed here is developed based on experiments carried out on the digital scholarly edition *Berlin Intellectuals*, for which *alterLDA* achieves high performance in the recognition of alterations on labelled data. On unlabelled data, applying *alterLDA* leads to interesting new insights into the alteration behavior of authors, editors and other manuscript contributors, as well as insights into sensitive topics in the correspondence of Berlin intellectuals around 1800. In addition to the findings based on the digital scholarly edition *Berlin Intellectuals*, we present a general framework for the analysis of text genesis that can be used in the context of other digital resources representing document variants. To that end, we present in detail the methodological steps that are to be followed in order to achieve such results, giving thereby a prime example of an Machine Learning application in the Digital Humanities.

## Editorial prolegomenon

Classical scholarly editing has a long-standing tradition in distinguishing between different types of editions (Witkowski, 1924). The characteristics of specific edition forms usually align with the intended readership, but they also take into account a bibliographic history that tends to differentiate more and more along time according to linguistic areas. In the German-speaking area, historical-critical editions that comprise an extensive historical-critical apparatus are often distinguished - with a clear hierarchical difference - from so-called study editions (Plachta, 2006). The common denominator between these two types of editions is that they aim to offer a "reliable text" as a central component (Plachta, 2006). In contrast to these types of editions, it is also possible to publish a reproduction of the manuscript image (facsimile edition). Plachta points out, however, that a facsimile edition is no substitute for the above two types of editions (Plachta, 2006).

Another way of differentiating between types of editions is to compare the intention in the text construction, which corresponds to the philosophy according to which the anglo-saxon area has mainly structured their approach. According to Andrews, “the ‘old’ methods that have their root in classical philology” strive to assemble the “ideal” text, while the “new philology” seeks to find the “real” text (Andrews, 2013). In this conception, the ideal text tries to approach the author's intention, while the real text seeks to emulate the existing sources.

The type of edition an editor goes for is often defined by economic factors in printed editions, while in digital editions, this limitation can be obsolete in terms of the amount of pages available, or located on a different level (for instance due to the price of specific, cost-expensive technologies). More generally, in digital scholarly editions, differentiation characteristics can be renegotiated. As Andrews states, there are hardly any technical limitations in digital editions with regard to the size of the apparatus, and the number and resolution of facsimiles provided (Andrews, 2013).

This is not the only specificity that distinguishes digital from print editions. They also are machine-readable. With digital editions being available in digital formats, computers can not only handle repetitive tasks in the creation of the edition (Andrews, 2013), they can also be used to perform tasks that use the edition as source material. The most obvious example for this type of use is the full-text search, but the machine-readable form also allows the creation of a multitude of statistics and customed visualizations with very little effort (Ralle, 2016). Furthermore, Ralle emphasizes that the digitization of editions and scholarly editing in general allow to pay special attention to the processual aspects of the edition (Ralle, 2016). An edition can be extended or enriched after it has been initially published and does not need to be ”finished” at a specific moment in time. A digital edition can be modified dynamically, for instance like the Carl-Maria-von-Weber-Gesamtausgabe with a front page field called ”What happened today?” that connects to all instances of the current date in the corpus and highlights them - a content that changes from day to day and offers a different approach to the corpus than the traditional keyword search. Also, user interaction can be funneled back into the edition, for example when subsequent publications that are based on the edition are listed there. Interaction in and of itself can also be included: the search behaviour of users can be analyzed for better future suggestions or the edition can be enriched by third-party data. Every user of a digital edition, whether computer or human, is thus potentially able to engage in one form of editorial participation or the other (Schlitz, 2014; Siemens, et al., 2012; Shillingsburg, 2013).

These special features of digital editions allow for paleography (Baillot & Schnöpf, 2015) to reach out to research questions hence unexplored in the Humanities due to the lack of tools and corpora allowing an automatic evaluation of alteration phenomena. It enables for instance to thoroughly reconstruct the history of a document by considering physical traces of alterations, meaning any smaller or larger text modifications on the manuscript, performed either by the author himself or herself or by others (see Methods). This approach provides insights into the way in which authors, editors and other contributors work together, hence impacting our understanding of text genesis as a collaborative process.

In order to achieve substantial results in this field of research, fast and well-structured access to the document variants is required. Digital editions presenting the manuscript alterations allow to focus on diplomatic transcription or facsimile, as opposed to print editions where the focus is on a single copy text, itself usually optimized for readability. Examples of digital editions representing the document history include faustedition.net (Goethe, 2017), bovary.fr (Leclerc, 2009) beckettarchive.org (Beckett, 2011-2018) and the edition that provided the background for the methodology we propose here: the digital scholarly edition “Letters and texts. Intellectual Berlin around 1800”, berlinerintellektuelle.eu (Baillot), *BI* in the following.

# INTRODUCTION

*Letters and Texts. Intellectual Berlin around 1800* is a digital scholarly edition of manuscripts by men and women writers of the late 18th and early 19th century. The connection these writers have to the intellectual networks in the Prussian capital city are either direct (authors living and writing in Berlin) or indirect (editorial or epistolar relationship with Berlin-based intellectuals). The originality of this digital scholarly edition is that it is neither author-centered nor genre-based, but presents different types of selected manuscripts that shed light on the intellectual activity of Berlin at the turn of the 18th to the 19th century. This editorial choice is presented at length in (Baillot & Busch, 2014), where light is also shed on the uniqueness of the Prussian Capital City in the context of the period. While correspondences play a key role in the edition, they are considered as a part of the circulation of ideas that is at the core of the project, so that letters, and more generally egodocuments, are complemented by drafts of either literary works (among which two major romantic texts), scholarly writings (one dissertation) or administrative documents (related to the development of the Berlin University). A first editorial phase (2010-2016) allowed to publish manuscripts that cover different thematic areas and historical phases of the development of intellectual Berlin. They were selected based mainly on their scholarly relevance and on their accessibility (the publication policy of archives holding manuscripts having a major impact on their integration to a digital edition that displays a facsimile like this one does). Four main topics have emerged in this first phase: French, e.g. Huguenot Culture, Berlin University, Literary Romanticism and Women Writers. Depending on the topics, the letters published were complemented by other types of texts that document the circulation of ideas and of literary and scholarly works in the late 18th and early 19th century.

The edition can be browsed by theme, by author, by period, by holding institution, or by date. The single document can be displayed on one or two columns presenting at the user's choice a facsimile of the current manuscript page, a diplomatic transcription, a reading version, the metadata, the entities occurring on the page and the XML file corresponding to the document. In this first development phase, 248 documents and 17 authors were encoded and presented in *BI*. In Figure 1, a quantitative summary of the BI corpus is given, which consists of introductory figures for the whole corpus in terms of size, temporal span and number of alterations and detailed information about individual authors.

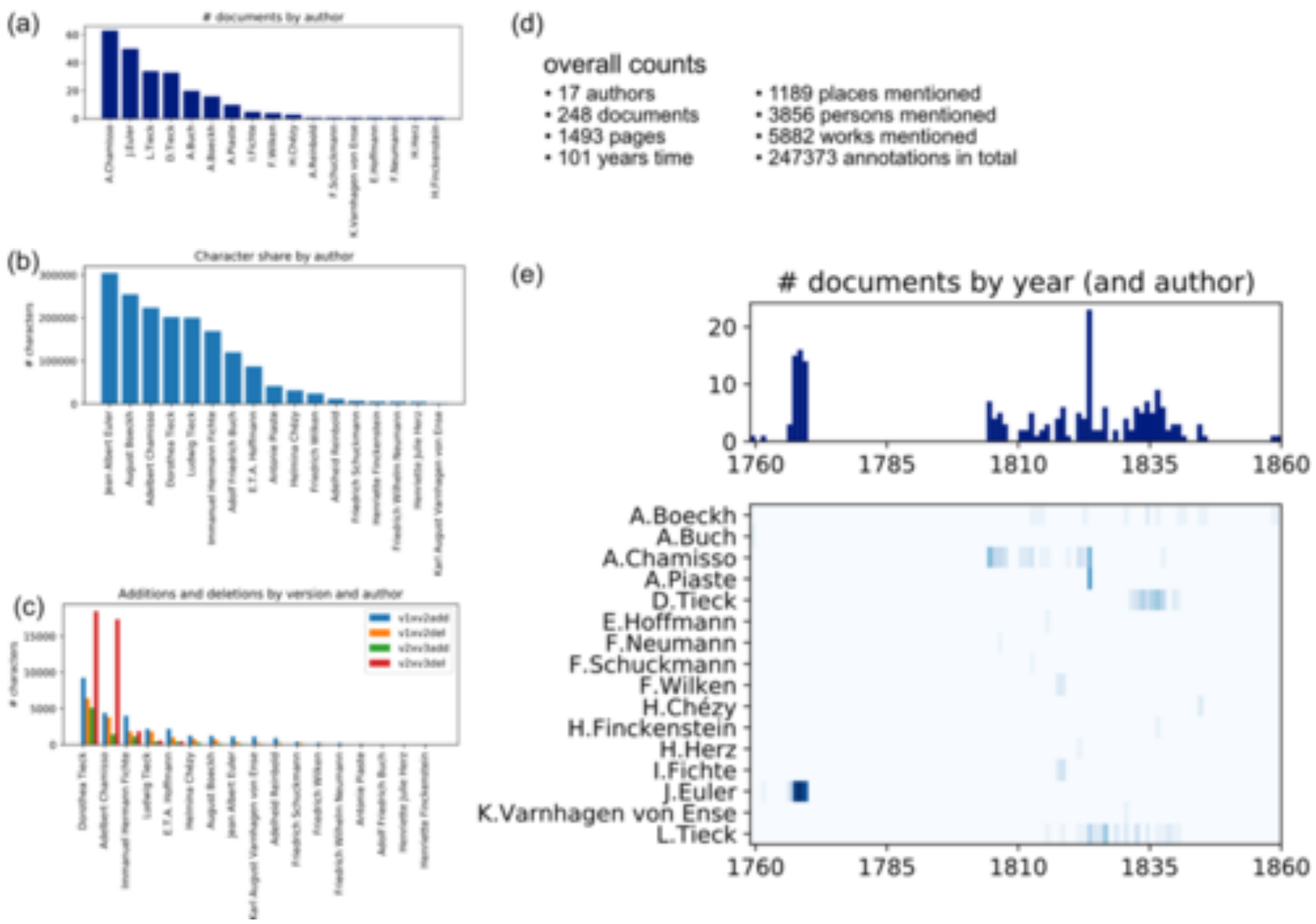

*Figure 1: Quantitative summary of the BI corpus. Subfigure (a) shows the number of documents grouped by each author. Adelbert Chamisso holds the largest share of documents. In (b), the actual number of characters by author shows a slightly different picture, instead of Chamisso, J. Euler has the largest portion. In Subfigure (c), the alterations in terms of "number of changed characters" between different versions of the documents are depicted for each author. A Large number of total characters of an author does not necessarily mean a lot of alterations. Although Euler encompasses the largest number of characters there are close to no alterations in the documents he authored. In (d), total counts for the whole BI corpus are stated to show the extent of the annotational effort. In (e), the temporal distribution of the creation of the documents is shown for the whole corpus (top) and for each individual author (bottom). In this subfigure ((e), bottom), the number of documents created in each year is encoded in the intensity of the color.*

A major novelty about the BI edition is that it combines genetic edition and entity annotation in order to gain insight in intellectual networks, on the actual editing process of manuscripts (of literary and scholarly works) and on the discourse about this editing process (letters – most letters are interestingly also partly transformed into literary works in their own right and subject to editing). The genetic encoding gives precise information regarding deletions and additions in the manuscript text. The BI encoding guidelines make extensive use of the following specific sections of the the TEI (P5)[f] guidelines additionally to the standard structure (Chapter 1-4): Manuscript Description (Chapter 10, https://www.tei-c.org/release/doc/tei-p5-doc/en/html/MS.html), Representation of Primary Sources (Chapter 11, https://www.tei-c.org/release/doc/tei-p5-doc/en/html/PH.html) and Names, Dates, People, and Places (Chapter 13, https://www.tei-c.org/release/doc/tei-p5-doc/en/html/ND.html), which offers the possibility to markup text alterations with tags such as "add" and "del".

As already mentioned, the BI Edition features annotations on the genesis of the documents (genetic edition), which, being a digital edition, are machine-readable. The core question we will address in the following is therefore whether machine learning models (Wainwright & Jordan, 2008) (Rasmussen & Williams, 2006) (Nakajima, et al., 2019) that analyze the alterations within the documents can be used to gain new insights into author, editor, and archivist practices, as well as practices of the intellectual societies in the document's creation time. The investigation of this question is only made possible by the meticulous (digital) annotation of the historical documents that provides previously unavailable enrichments and perspectives on the sources.

From the perspective of edition theory Ehrmann stresses that the importance of analyzing the alterations in manuscripts for literary studies and scholarly editing lies not only in the fact that they allow an insight into the author's writing process in the case of author-made changes, but also in the fact that they help identify the respective contribution in the case of co-authorships (Ehrmann, 2016). The first question that arises when examining every alteration is the question of the underlying reason, be it for a minor correction of mistakes or a wide-ranging content-related alteration. This leads to the question of the originator of the alteration and, as Ehrmann stresses, whether the alteration is wanted by the author (Ehrmann, 2016). In the specific case of an edition of correspondence, the intended readership is bound to change dramatically in the aftermath of publication. A letter that was originally written to a friend is made public to a large readership and in the process of preparation, the editor applies alterations to the original letter, most of the time with a correction phase on the original manuscript itself. This is the case for many manuscripts in the BI edition, commented on as follows by the editors:

*One characteristic of letters is that you generally are not the first one to read them when you discover them in an archive. Not only have they been addressed to a person or a group of persons in the first place [..], many of the letters we at least are working on have already been edited in the last centuries. But not in extenso, no: they have been abridged, overwritten, corrected according to the expectation of the audience in the time that they were edited. (Baillot & Busch, 2015)*

Moreover, the novel machine learning method (alterLDA) presented here also offers new opportunities for many other areas of automated analysis of variants of sources, especially within the Digital Humanities. AlterLDA is based on the topic model latent Dirichlet allocation (LDA). "Topic Modeling has proven immensely popular in Digital Humanities" (Schöch, 2017). LDA is particularly popular in the DH because it is suitable for explorative text analysis. With the automated compilation of word lists by LDA, new topics can be identified in large text corpora whose existence was previously unknown. In this context, it almost always forms the first analysis step on text data, but it can in fact also be used for non-textual data (Jelodar, et al., 2019) (Liu, et al., 2016). In addition to LDA, which provides the identification of the overall relevant topics of the corpus to be examined and the specific topics of the individual documents of the corpus, alterLDA is particularly concerned with the variants of the documents. The starting conditions for this work are as follows: from the point of view of edition theory, the question of document variants is of major importance, and this has not yet been sufficiently investigated with Distant Reading methods. From a methodological point of view, there is a very widespread Topic Model (LDA), which is already recognized and accepted practice in the Digital Humanities. In this paper we therefore close the gap by adapting LDA in order to model document variants.

The processual aspects of text genesis in the sources underlying the edition are thus highlighted and supported by the processual aspects of the edition itself. If a document in its past has already been prepared for publication by an editor, then his or her notes in the TEI-XML are annotated in the same way as when the editors of the BI Edition leave notes: with the <note>-tag.

Parts may be deemed inappropriate for publishing to a broader readership at a certain place and time due to their political or religious context, or for revealing private information about a person or a group.

The application on the BI corpus is particularly interesting because the latter consists mainly of letters, which, especially around 1800 in Germany, exhibit a strong tension between public and private sphere. The framework presented here includes four methods that range from basic, well established, rule-based methods to a specialized, novel machine learning method (alterLDA) that was developed for exactly this purpose. From a methodological point of view, this is a challenge for all disciplines involved, conceived as a scenario optimized so that all sides benefit from each other. Finally, the newly introduced method is also applied to discover alteration candidates in the documents that are not yet altered. These findings led to, and hopefully will continue to fuel, interesting discussions on parts of the edition that were unnoticed thus far.

Due to machine readability, documents in digital editions can be modified by computer programs in such an algorithmic way that they are transformed into something else. This transformation is described by Stephen Ramsay in the concept of Algorithmic Criticism:

*Any reading of a text that is not a recapitulation of that text relies on a heuristic of radical transformation. The critic who endeavors to put forth a "reading" puts forth not the text, but a new text in which the data has been paraphrased, elaborated, selected, truncated, and transduced (Ramsay, 2011, p. 16).*

The methods to which Ramsay refers here, e.g. tf-idf normalization, are mostly deterministic methods. In this work, however, a probabilistic method is used to transform the documents and their variants, which uses previously collected data and relates the observations to it. Thus, these statistical transformations resemble human reading more than purely deterministic approaches and therefore foster the methodological concept of the “radical transformation” described within the Algorithmic Criticism towards a more general criticsm that includes non-explicit algorithms.

This transformation in an algorithmic way is a very standard technique (e.g. counting co-appearances of speakers in scenes of a play). However in recent years Machine Learning models are being used more broadly for e.g. Named Entity Recognition (Dalen-Oskam, 2016) (Jannidis, et al., 2015). The patterns that are used to identify entities are not stated explicitly by a programmer but are learned from the data at training time. In general, Machine Learning methods would then be methodologically less strict than classical explicit algorithmic transformations and therefore be possibly also more human-like. However, when giving up this explicitness there has to be a more rigorous evaluation of the machine’s output. For many applications, like the one presented in this work, we therefore rely on the evaluation by machine learning and humanities scholars, who can employ methods for interpreting and explaining machine learning models (Samek, et al., 2019).

# METHODS

In this section, the machine learning methods for identifying the reason for a given alteration are presented, by first introducing the general data analysis pipeline. Then, we specify precise definitions of the most relevant concepts for alterations. After specifying the preprocessing steps, the novel alterLDA model is introduced. It is designed to analyze the most interesting, yet most complex types of alterations. Before the methodologies of each step are explained in further detail, the definitions for the most important and most frequently used terms are given here.

Given an arbitrary version of a document, we define an alteration to be a local group of added and/or deleted symbols that is performed by the author of an alteration. Basically, any symbol appearing in the document could be regarded as a single addition, but the state of the manuscript at the time of the investigation often makes it impossible to identify beyond doubt which groups of symbols belong to a particular writing session. The same problem exists with deletions: Was the sentence completed first, or did the author pause in the middle and correct something before completing the sentence? In BI, additions and deletions are considered as such when they clearly stand out, for example when they are crossed out or written to the margin. Sometimes co-occurring additions and deletions are also referred to as replacements. The alteration may range from a single character to whole passages of the document and can even be a local group with non-altered symbols in between. An alteration author is a single person or institution that alters the document, possibly the primary author him or herself. The alteration author has an alteration reason for which he or she decides to alter the document. This is a very specific reason, for example "the alteration author thinks that a particular word is spelled differently" or "a real person which is referred to in the document may not want to be recognized by the readers, so this part is censored".

Each alteration has a formal and content-related portion with varying emphasis. For example, if the author of an alteration changes the spelling of a single word this would not change the meaning of the document in most cases. On the contrary, adding multiple sentences to a document may change the content of the document significantly. Of course, whether an alteration is rather positioned on the form side or on the content side of the axis depends on the point of view of the recipient. Hence, the proposed method takes into account the formal changes of the document as well as the content-related changes. Smaller alterations tend to have a rather formal aspect, where longer alterations almost always are content-related.

The set of alterations can be broken down into different categories with respect to their alteration reason. One group of alterations is (1) the group of paratexts, for example archival notes, such as numberings or dates, or stamps and seals of the library or archival institution. Another group of alterations is (2) corrections of mistakes which consists in spelling alterations, grammatical changes and other corrections. (3) The third group contains stylistic alterations, for example replacing a token with its synonym or rearranging the word order. Of course, changing the word order is sometimes more than just a stylistic change, but one could e.g. begin a sentence with "Es bedarf daher [..]" as well as with "Daher bedarf es [..]" with very similar intentions. The last group of alterations which we call (4) content-related alterations incorporate alterations that either add new information to the document or suppress information that was present in the document before.

*Figure is not included in the preprint as the permission to publish the image sources only covers the journal article*

*Figure 2: Flow chart of machine learning pipeline with four example alterations. The stream of documents is analysed in four steps that identify different reasons of alterations as depicted in the panel at the top. In the panel at the bottom, the details of the individal alterations are presented. Each alteration has a unique appearance and unique characteristics, like the type of ink and the way in which it fits into the surrounding script. The presented preview of the facsimiles are shown in greater detail in Figure 9, Figure 10, Figure 11 and Figure 12.*

Figure 2 illustrates how the identification method works. All alterations are put into the analysis pipeline, and after the initial distinction between author alterations and non-author alterations, the four tests for different types are performed on each alteration. As an example, there are four alterations depicted in the illustration that are fed into the model. A detailed explanation for an identification of the three non-content-related types of modifications is given in the appendix. By elimination of all other possible categories, the remaining alterations are of the content-related category. There are still a variety of reasons in this category worthwhile to identify. Rather than the general category we aim for providing a distinct reason for each alteration. The fourth alteration which is marked in red is a longer deletion and a detailed facsimile is shown in Figure 2. It is performed with a pencil which is different from the primary ink of the letter. It deals with the author's sickness and with the sickness of the author's mother. The extent of the alteration already indicates that this is not a correction of a mistake and since the part that is deleted is not replaced by anything else, it can be assumed that this alteration changes the amount of information provided. It is thus to be classified as a content-related alteration. At this point it is still to be identified for which specific reason the document has been altered. With alterLDA, the alteration is assigned to one of a set of candidate reasons as a final step, in this case *Sickness*-reason.

## Related Work

We convey a generative topic model, that is based on Latent Dirichlet Allocation (Blei, et al., 2003) and that is able to take into account the structural information of alterations. LDA is a widely used topic model that extends Latent Semantic Indexing (Deerwester, et al., 1990) which is capable of assigning a distribution of topics to a document instead of only a single topic. LDA takes advantage of the fact that a text is organized in documents. This structural information is the reason for LDA to function. Based on this structure of documents, LDA can learn which words tend to co-occur and thus have a relation. Words that often occur with each other form a topic. In this context, a topic is merely a distribution of word frequencies.

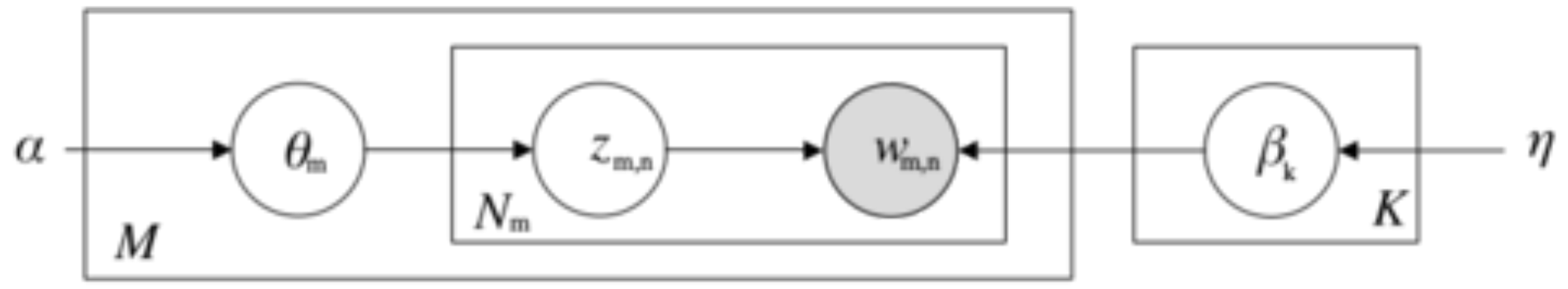

*Figure 3: Graphical representation of the LDA model. The plate notation visualizes the generative process of a probabilistic model by following the directions of the arrows. Given $\alpha$ and $\eta$, one initially draws $\beta$, a distribution over words for each topic and $\theta$, a topic distribution for each document. Then, for each token within a document, one draws a topic assignment and only then (because $\mathrm{w}$ has input arrows from both, $\beta$ and $\mathrm{z}$, one can draw $\mathrm{w}$ from the topic in $\beta$, that was assigned in $\mathrm{z}$.*

In Figure 3, the LDA model is shown in plate notation. An overview over the used symbols is given in the appendix. The plate notation shows the graphical representation of the LDA model. An open circle denotes a model variable and a shaded circle denotes an observed variable. Symbols without circle denote a hyper parameter. A rectangle indicates repetitions of the included variables. In this model, $\beta$ represents the topic histograms, $\theta$ represents the topic mixture for each document, z represents the topic assignment for each token position and **w** denotes the token itself. LDA has no notion of the order of words within a document, which is referred to in the literature as a "bag-of-words" for each document.

There exists a wide range of topic models that customize LDA by taking into account additional structural information. To replace the bag-of-words approach by introducing structural information about the word order is a major field of LDA research (Rosen-Zvi, et al., 2004; Gruber, et al., 2007; Wallach, 2006). In addition, there is a broad research community that addresses the recognition and arrangement of hierarchies of topics (Blei, et al., 2010; Paisley, et al., 2015). LDA has also been modified to work with graph-structured documents (Xuan, et al., 2015). However, we are not aware of any literature that shows how to model alteration reasons in a corpus of natural language. Therefore, this paper is an important contribution to close this gap, i.e. to provide the literary scholarly community with a novel method and to open up another field of application for the machine learning community.

## Alteration Latent Dirichlet Allocation

In Figure 4, the alterLDA model is described in plate notation. The upper part is standard LDA whereas the lower right part contains the newly introduced variables to model alterations.

In standard LDA, the observed variable (the input) is just the words within each document. In the alterLDA setting, the additional structural information about the alteration of each word is provided as input. With that, the alterLDA model tries to infer the tendency for each topic to be an alteration topic.

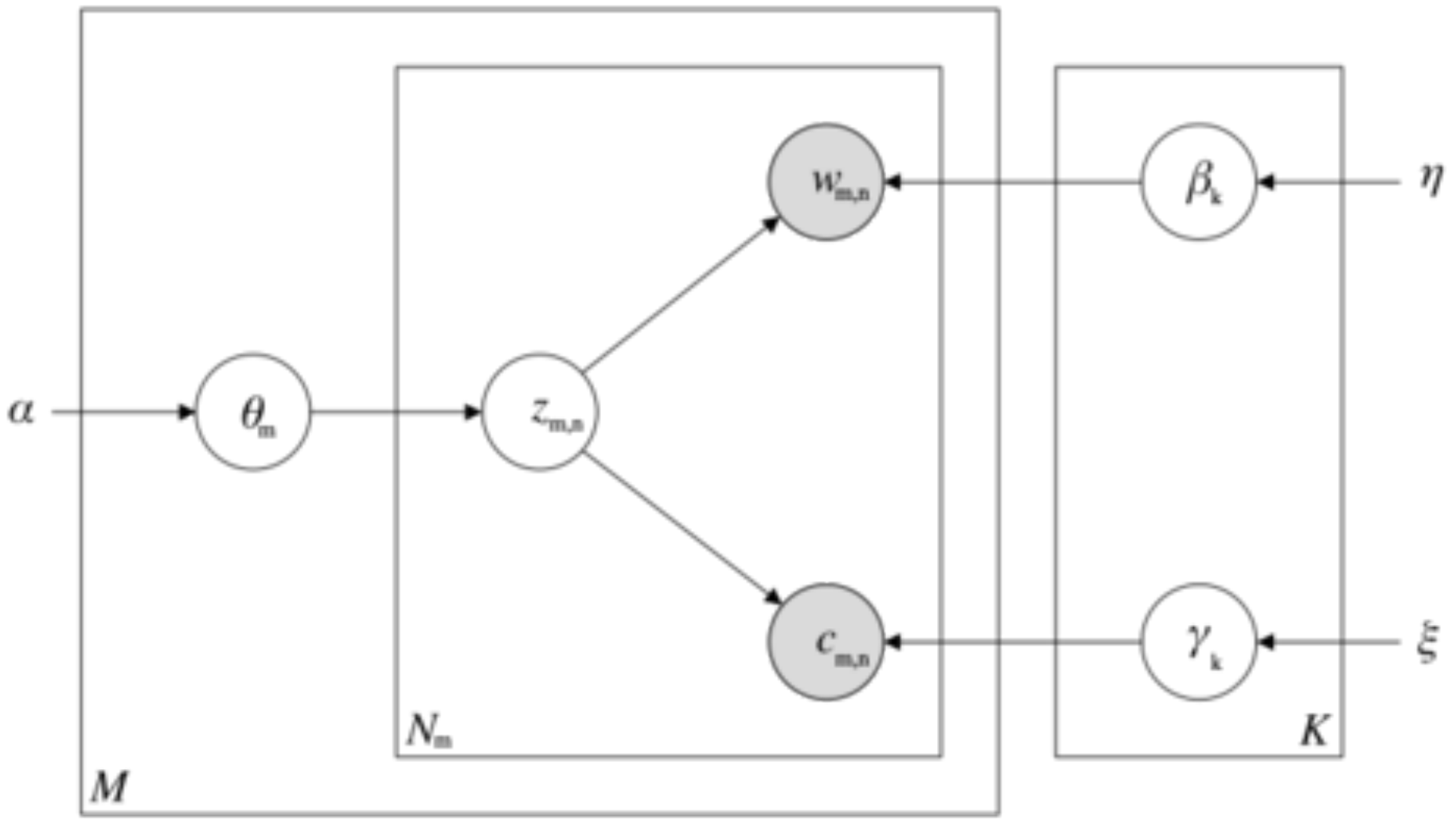

*Figure 4: Plate notation of the new alterLDA generative model. Newly introduced is the lower branch with variables* ***c****, γ and ξ that deal with alterations. There exist M documents with $N_m$ tokens each, Also, there exist K topics and for each topic, there exist a tendency for it to be a reason for alteration (γ).*

The generative model detects reasons by taking into account all text, inside and outside the alterations. From alterations that were gone through manually, we expect to see alteration suggestions that mainly relate to the privacy of a person, political or religious topics may appear as well. In order to make the model description as clear as possible, we try to keep the mathematical formulations to a minimum. Therefore, we only include an explanation of the symbols used (see appendix), a graphical representation and the derivation of how the model can be algorithmically captured using a Collapsed Gibbs Sampler. Similar to LDA, in alterLDA there exists no feasible algorithm to compute the posterior distribution of the latent variables. Instead, approximate methods need to be applied to find a solution in reasonable time.

A Collapsed Gibbs Sampler is one of the possible approaches to find an approximate solution to the objective. Generally, a Gibbs Sampler iteratively samples the configuration of a specific latent variable based on the current configuration of all other model variables. An introductory tutorial on Gibbs Sampling LDA is presented by (Carpenter, 2010). This algorithm can also be understood as an instance of a Markov Chain, a constrained iterative probabilistic model itself, where the current state only depends on the previous. From this perspective, the stationary of the Markov Chain represents the solution of the Gibbs Sampler. The source code of our implementation of the Collapsed Gibbs Sampler for the alterLDA model is publicly available.[g] In the appendix, derivation of the Collapsed Gibbs Sampler for the alterLDA model is given.

## RESULTS

In this section, we present three experiment settings which mainly differ in the splitting between training and test data. As shown in Figure 5, three settings are chosen, S1 as a straightforward explorative demonstration, S3 to comply with the methodological standards of data splitting for the performance report, as well as S2 for offering additional explorative results specific for this data set. We will first present the evaluation results that investigate the performance of alterLDA on the given data set and afterwards present explorative results that will be reconciled with expert knowledge. Apart from these experiments on the BI data set, the first experiments were performed on synthetic data, some results from these experiments are listed in the appendix.

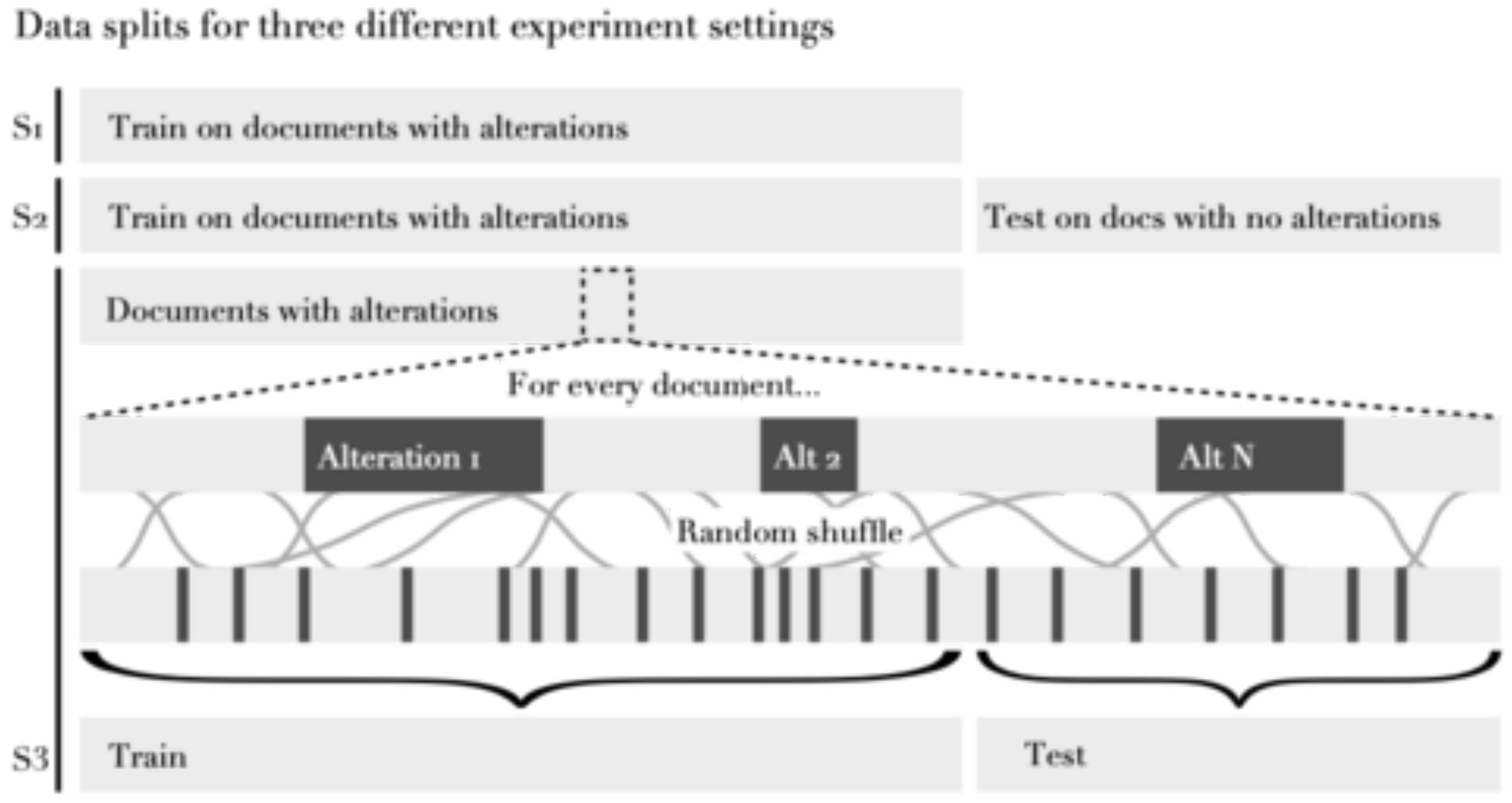

*Figure 5: Visualization of the data splitting setup for all settings. For each experiment a different data setting is used. The different Settings are shown in the leftmost column (S1, S2, S3). Within each setting, the row at the bottom depicts the final setting of the data. Setting 1 (only one row) does not require test sets, Setting 2 (only one row) aims at finding alteration candidates in texts with no alterations. For Setting 3, the process of creating the setting is depicted in multiple rows. First, only documents that contain alterations are chosen. Then, each individual document is shuffled and split into a training and a test part.*

### PERFORMANCE EVALUATION

In this experiment, in which alterLDA is applied to the entire training data, it is to be determined whether the model in principle delivers plausible results. It will be verified whether $\gamma$ finds a meaningful topic composition that represents sensitive topics. This means that alteration topics may be a convolution of private and maybe political and religious matters.

With alterLDA, various parameters must be set which influence the outcome. These are the same parameters as for LDA: Number of topics and the Dirichlet prior for the topic distribution $\eta$ and the topic mixture $\alpha$. There is also another parameter, the Dirichlet prior for the alteration tendency $\xi$. The default value for a Dirichlet prior is 1, but it can take any value greater than 0. The smaller the value, the more the variable tends to be focussed on single values, the larger the value, the more different values are considered. Using the topic mixture as an example, a small $\alpha$ would mean that LDA is looking for a solution where each document consists of only a few topics, a large $\alpha$ finds a solution with mixtures of many topics. AlterLDA is initialized in this setting with $\alpha$

= (.1, .1, …), $\eta$ = (.1, .1, …) and $\xi$ = (.05, .05, …) as well as K=10. We choose $\xi$ small in this setting to create a sparse $\gamma$ so that alterLDA only learns one alteration topic.

The resulting topic learned in this naive approach as alteration-sensitive is visualized as a word cloud in Figure 6. It is very difficult to put a single label on this "topic". The most probable words are strongly influenced by global word frequencies, the strongest four words describe it: "Sie Ich Brief schreiben" this does not come as a surprise since the corpus consists mainly of letters. However, it is also possible to find terms from any subject area that was suspected in advance of being altered: *Sickness* terms are for example "Operation", "Bett", "fürchten", *Financial* terms are e.g. "Geschäft", "Geld" and regarding *Love Story* there is for example "lieb" and "schön".

Beforehand, we assumed to also find political, religious topics but these do not appear in the naive setting whereas diverse private topics do occur.

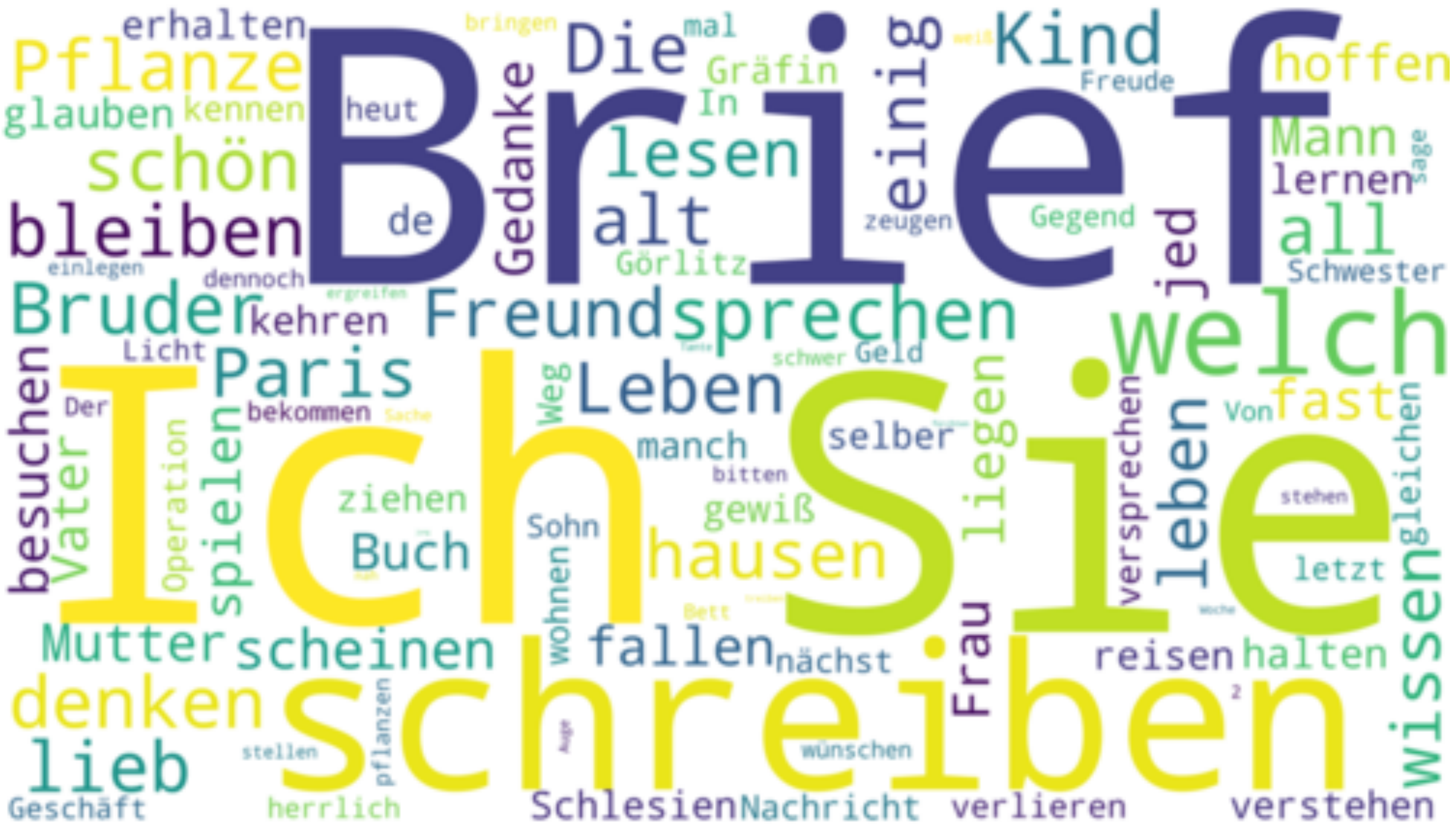

*Figure 6: The strongest words of the topic that has a high alteration tendency after training alterLDA in the naive setting (Setting 1). The stronger the word, the larger the font. Strongest words are very general words on the topic of letters, words from the Financial, Sickness and Love Story are also weakly present.*

As visualized in Figure 5, the BI corpus consists of documents with alterations and documents without alterations. If we want to measure the performance of alterLDA in predicting the tendency for alteration, documents with alterations are much more helpful.

In Setting 3, we only use documents with alterations to produce the training and test set. We split every document individually into training and test set after shuffling to increase the chance that alterations are present in both sets (Müller, et al., 2001). After training, we use the topic mixture $\theta$ of the corresponding document.

In this setting, alterLDA is initialized with $\alpha$= (1, 1, …), $\eta$ = (1, 1, …) and $\xi$= (1, 1, …) as well as K=20. We explicitly chose the Dirichlet priors all equal to 1 as this can be considered the default. To allow for more topic diversity, we chose the number of topics a little bit higher than in the naive setting. This parameter combination will be used throughout the rest of the paper.

The performances on the total test set as well as for each individual author are shown in Table 1. The performance varies considerably across different authors where D. Tieck, L. Tieck and Ad. Chamisso work well above chance level, the performance for Hoffmann and especially H. Finckenstein is weaker. In case of H. Finckenstein, this may be due to the fact that in the corpus there is only a single letter. For E.T.A. Hoffmann, there is also only one document in the corpus, but it presents two specialties. It is considerably longer than most documents in the corpus: it is not a letter, but the novella *Der Sandmann*. The larger size and the differing properties due to the genre seem to trade off to a slightly better performance than in the case of H. Finckenstein. We thus argue that the performance of alterLDA depends on the size of the training set and on the homogeneity of the documents.

The results of this setting are not meant produce new domain insights as it only aims at reproducing the alteration tendencies of already altered documents. However, screening performance difference across viewpoints such as *authors* still reveals properties of the underlying data set.

| *Grouping* | *Balanced Accuracy* | *Area under ROC* |
|---|---|---|
| *Adelbert von Chamisso* | .60 | .57 |
| *Henriette v. Finckenstein* | .38 | .07 |
| *Immanuel v. Fichte* | .49 | .64 |
| *E.T.A. Hoffmann* | .5 | .53 |
| *Dorothea Tieck* | .61 | .65 |
| *Ludwig Tieck* | .69 | .65 |
| *Total* | *.67* | *.66* |

*Table 1: Test set performance for documents with alterations. The test set is grouped by author and two performance measures are given. Balanced Accuracy and Area under Receiver Operating Characteristic. In both cases, the sklearn implementation is used. http://scikit-learn.org/stable/modules/classes.html#sklearn-metrics-metrics version 0.20.0.*

## Explorative Analysis

In the explorative experiment (S2), the corpus is divided into two parts: On the one hand, all documents that contain changes and, on the other hand, all documents that do not contain any changes. The aim of the experiment is to train the model on the part of the corpus that contains changes and then let the model suggest which parts of the unchanged corpus may be changed in a similar way. There may be different reasons why some documents contain alterations and others do not. Assuming that all documents were reviewed by the same person and that person was also so diligent that he or she did not overlook a single passage, then alterLDA should at best-case scenario not propose an additional passage to be altered. We assume in this experiment that either not all documents have been reviewed for the same criteria or that relevant positions have been overlooked.

| *Author* | *Suggested alterations* |
|---|---|
| *Immanuel Hermann von Fichte* | 3 |
| *Karl August Varnhagen von Ense* | 16 |
| *Friedrich Wilhelm Neumann* | 54 |
| *Helmina von Chézy* | 59 |
| *Adelheid Reinbold* | 73 |
| *Henriette Herz* | 76 |
| *Friedrich von Schuckmann* | 118 |
| *Antonie von Chamisso* | 258 |
| *Friedrich Wilken* | 340 |
| *Ludwig Tieck* | 389 |
| *August Boeckh* | 540 |
| *Adolf Friedrich von Buch* | 907 |
| *Dorothea Tieck* | 929 |
| *Adelbert von Chamisso* | 1075 |
| *Jean Albert Euler* | 2558 |

*Table 2: The table shows the number of suggested alterations for different authors/editors. Only documents that were not truly altered are included. The number of suggested alterations represents the number of positions in dochuments that the method suggests to alter. Euler and Buch mainly wrote in French, which influenced the prediction a lot for these authors. Chamisso wrote in German although his mother tongue is French. Therefore there may be many minor mistakes that are suggested to be altered. Dorothea Tieck wrote about her mothers' sickness which the method recognized as a sensible topic and therefore as a reason for alteration.*

In Table 2 the counts of the positions in documents with no alterations that have been suggested by the method are displayed for each author/editor. The rows in the table are sorted by the total amount of suggested alterations. Interestingly, the authors with many suggested alterations are not necessarily the ones that have a large share of total tokens of the corpus (see Figure 1, (b) and (c)). In the case of Euler and von Buch, this is due to the fact that their documents are mostly in French, whereas the alterLDA model in this case is primarily trained on German texts. For Boeckh, this is mainly due to the fact that the corpus encompasses only a few yet long documents and consequently there are not many documents present in the test set. Of course, there are other reasons for each author's ratio of corpus portion and number of suggested alterations. A person that altered all positions in the training set also diligently edited all documents in the test set and simply did not find any position that should be altered for the same reason: That the method did not find the respective amount in the test set can either mean that it was not able to find the right positions or that there were none.

For further analysis, we will ignore the texts by J. A. Euler and A. F. Buch and focus on the other four authors for which alterLDA suggested most alterations. As said, the texts by J. A. Euler and A. F. Buch were mainly written in French which influenced the number of suggested alterations. In Table 3, the most common words that were suggested to be altered for individual authors are listed. For all authors except A. Boeckh, the majority of words seem to relate to the overall *letter* topic, however for D. Tieck the keyword "Krankheit" *Sickness* appears. For Ad. Chamisso, words like e.g. "schön" and "begehren" can be observed that may relate to the topic *Love Story*. In the case of L. Tieck, a distinct convoluted alteration topic is not immediately conceivable. In the case of A. Boeckh, the topic of the documents in the corpus are mostly academia-related.

| *Author* | *25 most common suggested alteration words (descending order)* |
|---|---|
| *Dorothea Tieck* | Sie, Brief, schreiben, Ich, schön, gewiß, denken, einig, lesen, all, gleichen, Düsseldorf, Agnes, Krankheit, Freund, kennen, erhalten, Arbeit, Dresden, halten, weiß, Die, Leben, Berlin, Lüttichau |
| *Adelbert v. Chamisso* | Brief, schreiben, Ich, de, Die, weiß, kennen, all, wissen, schön, 4, Freund, denken, Sie, gleichen, halten, 3, neu, ton, erhalten, begehren, bleiben, einig, lesen, Sache |
| *Ludwig Tieck* | Sie, Freund, Ich, Geist, Brief, Tieck, Dresden, Von, Ihr, umarmen, Juli, halten, sogleich, erleben, Die, schwach, schweigen, sprechen, Mich, Herrn, Vergnügen, fordern, Masse, gleichen, eintreten |
| *August Boeck* | Mitglied, Seminar, Sie, Prämie, Fichte, erhalten, Arbeit, 1813, 2, Verfasser, 1812, Übung, hiesig, 4, Fähigkeit, welch, außerordentlich, Nummer, zahlen, Prüfung, Wernike, Anstalt, anfangen, Gedicht, Studiosus |

*Table 3: Most common words that were suggested to be altered in texts from individual authors.*

For a better understanding of alteration suggestions, a closer look into the individual authors is provided. D. Tiecks' documents reveal a sequence of letters that she wrote to F. Uechtritz in the years between 1831 and 1840. In the letters she repeatedly mentions her mother's sickness until her death in February 1837. Later, in March 1837, the father of F. Uechtritz passed away as well, D. Tieck writes about this in Letter 28.

By manually reviewing this series of letters, the editors of the BI edition agreed in many cases with the classification of the alterLDA model that D. Tieck's mother's sickness plays a role for the alteration tendencies of the documents. In some cases, however, human experts and the model disagreed about the reason for alteration. The following excerpt from letter 12 is identified by our proposed method as a stylistic alteration.

*Meine arme Mutter*

*~~leidet schon seit längerer Zeit an Unter=~~*

*~~leibsbeschwerden, der Arzt sagt es seyen~~*

*~~Verhärtungen und Anschwellungen der Drü=~~*

*~~sen, sie~~ hat schon seit längerer Zeit viel zu leiden, ~~braucht schon~~*

*~~seit 3 Monathen~~, trinkt seit 4 Wochen hier*

*Karlsbad, und ~~alles~~ bis jetzt ohne den*

*mindesten Erfolg.*

*(BI, Dorothea Tieck to v. Uechtritz, Letter 12, p. 2)*

And one can argue that this is actually a stylistic alteration, because the information about the mother's sickness is preserved after the alteration. However, the detail that her mother has pelvic complaints is suppressed in the second version - this discrepancy in detail can be decisive for classification as a content-related alteration.

In Figure 7, the number of suggested alterations (for the test set) and the number of actual alterations (for the training set) are displayed for each document by Ad. Chamisso. Most of the letters are addressed to L. La Foye (Ad. Chamisso's best friend), some are addressed to Antonie von Chamisso (Ad. Chamisso's wife). For each of the addressees, the letters are ordered by date. There are two letters (letter 10 and 11) which stand out significantly with regard to the number of alterations, letter 10, actually encompasses a large number of alterations, whereas letter 11 is part of the test set and thus does not have any (content-related) alterations. The alterLDA model suggests an almost equally high number of alterations for letter 11, presumably because it consists of topics that the alterLDA model estimates to be altered accordingly - this shows that the alterLDA model captures subtle changes of topics by the same author.

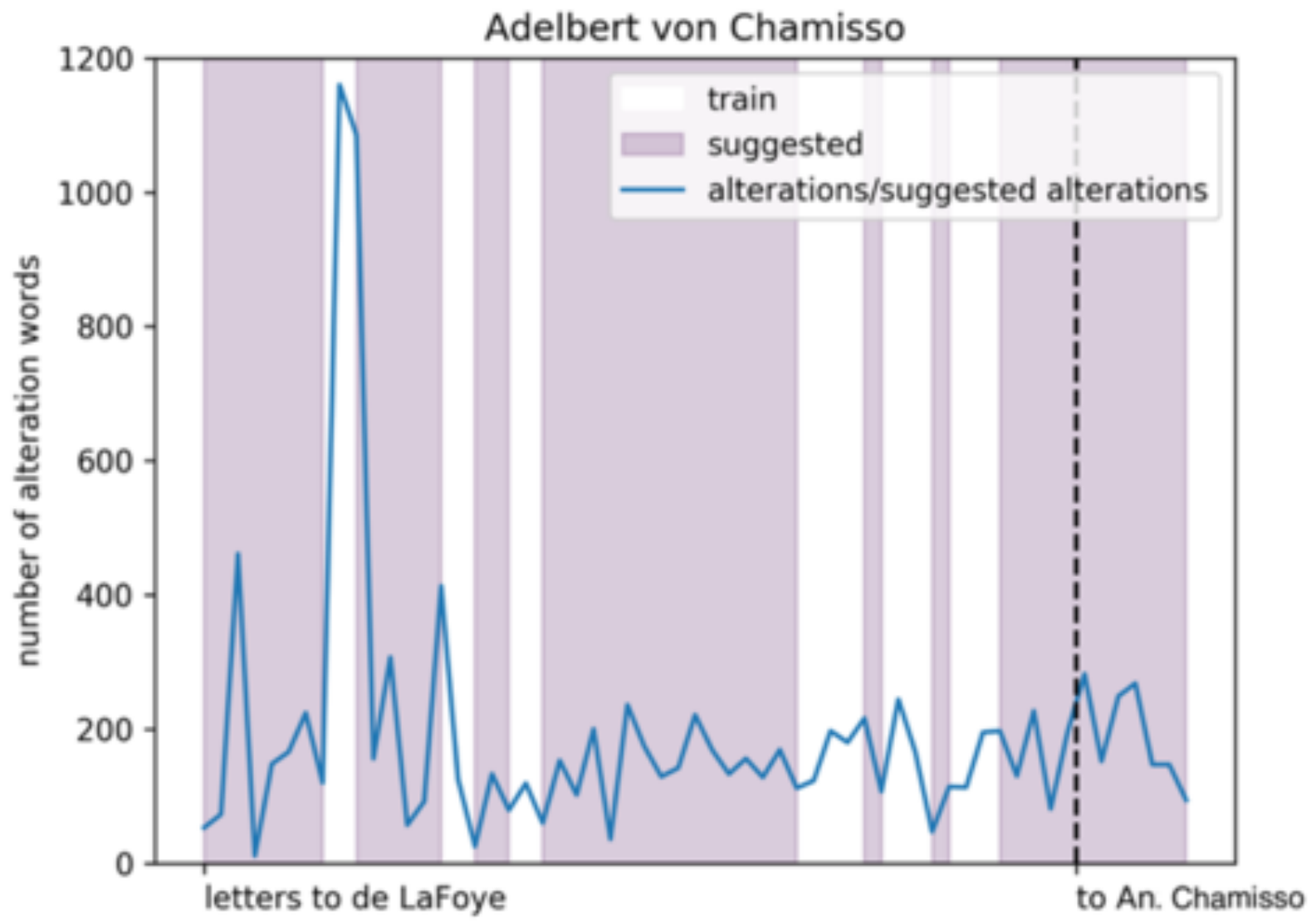

*Figure 7: The letters are divided into documents containing content-related alterations (white background) and documents without content-related alterations (purple background). The alterLDA model is trained on the white part and predicts possible alterations on the purple part, so the blue line shows the number of real alterations and suggested alterations, depending on the background. Left of the dotted separator, we find letters addressed to L. La Foye, on the right side letters addressed to An. Chamisso. There are two consecutive outliers with significantly higher numbers of alteration words. One is part of the training set, one is part of the test set, the temporal proximity may indicate a content-related proximity that the model was able to capture.*

In Figure 8, the correspondence from L. Tieck to F. Raumer that is depicted ranges from years 1815 to 1840. The left panel showing the number of letters that were sent during that year grouped by whether they contain content-related alterations. The right panel shows the number of tokens that were altered (blue) and the number of tokens that alterLDA suggests to be altered (orange).

Just comparing the blue bars of the two panels reveals that despite the fact that in 1836 there was only one letter written, there occurs the third-highest number of altered tokens. By examining the letter, it turns out that Tieck wrote about his financial problems and his plans to sell his book collection to the Count Yorck von Wartenburg.[h]

Referring back to Table 3, Financial terms are not present in the most common alteration suggestions for L. Tieck. This could indicate that the person editing L. Tieck's letters did not miss parts that refer to this financial struggle.

When also considering the suggestions by alterLDA (the orange bars of the panel on the right), one letter from 1838 draws the most attention just by the sheer number of suggested alterations. In this letter, L. Tieck refers to disputes between the Catholic Church and the Prussian state at that time.[i] By arguing about this political controversy, L. Tieck chooses his wording in such a way that alterLDA suggests alterations. This could indicate that across the training corpus there might be the same tendency to alter parts of the documents that deal with a political controversy. The fact that alterLDA highlights a document containing a mixture of political and religious topics supports the hypothesis that the alterations in the BI corpus do not only consist of privacy matters, but also involve a wider political dimension. This result confirms and gives a novel dimension to the assertion that letters as a text genre evolve, especially in the German context of the 1800s, at the interface between private and public matters. In that sense, the role played by alterations aiming at balancing private and public dimensions is central and needs to be further delved into. AlterLDA provides a systematic approach to this major issue in literary studies.

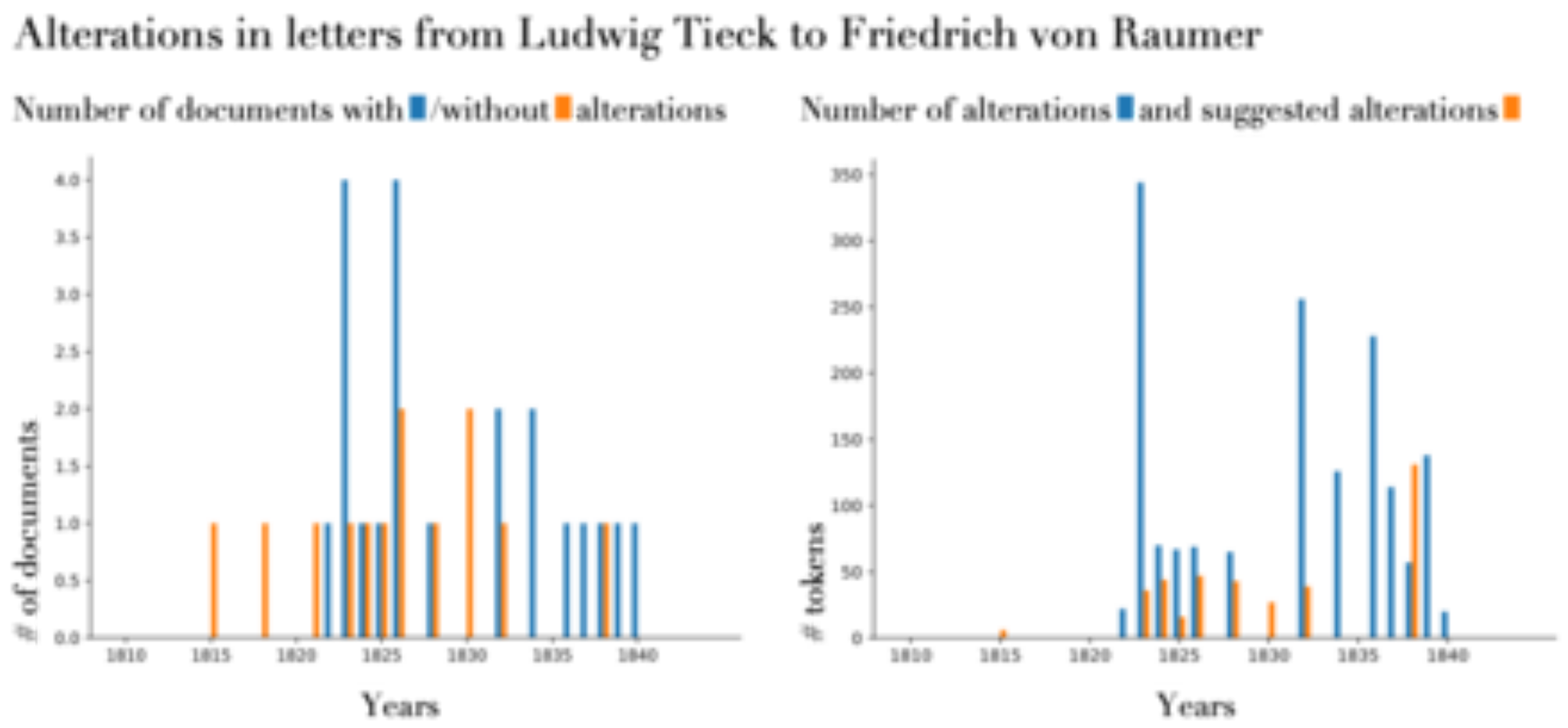

*Figure 8: Left panel shows the timeline with counts of letters from Ludwig Tieck to Friedrich von Raumer. The right panel shows the number of tokens that were altered (blue) and the number of suggested tokens (orange). The comparison between the number of letters and the number of alterations for each year shows that there are times where the letters were altered more (e.g. 1836). The letters with suggested alterations deal with financial, political and religous topics.*

## CONCLUSION

This paper presented a general framework for analyzing alterations in historical documents, ranging from simple error corrections to stylistic changes and even to content-related alterations. In addition to established methods such as regular expressions, string distances and vector space comparison, a new probabilistic model for the classification of reasons for alterations has been introduced (alterLDA).

This work contributes to the understanding of text genesis, as it provides insight into the layers of changes in documents. It also offers a quantitative way of evaluating which topics are at what times prone to be altered and are therefore sensitive.

From a machine learning point of view, the BI data set posed special challenges because, on the one hand, the data set is very small and, on the other hand, it comprises several languages, which also differ greatly from the ones used today. Nonetheless, alterLDA was able to confidently find alterations on unseen, labelled data. Exploratively, the method was able to find characteristics on unseen, unlabeled data that in many cases match the expert analysis. The method hence proves to be useful to draw the human reader's attention to specific parts of the large corpus that may otherwise be unnoticed, and by doing so serves as an example of how a machine learning method may assist a scholar as a collaborating reader and a potential collaborating editor.

With regard to the editorial issues first presented here, it is to be noted that the machine-readability of the BI edition makes it possible to serve problem-specific, individualised editions tailored to the research question of a reader/scholar. This project showcases how machine learning methods radically transform the way in which scholars engage historical documents, by taking advantage of the quality of deeply-annotated data: editorial and machine learning expertise can be brought together to explore in depth Humanities research questions.

This research benefited greatly from expert knowledge on the corpus as well as from novel ML methods that were designed for this corpus. This collaboration therefore presents important guidelines for practical and methodological steps that will help other projects to enrich their digital editions with automated annotation or ML-guided corpus exploration.

To achieve further progress, highly interdisciplinary research is mandatory where novel ML models are conceived, and domain knowledge from literary studies is interacting with statistical inference.

# BIBLIOGRAPHY

Andrews, T., 2013. The third way: philology and critical edition in the digital age. *Variants,* pp. 61-76.

Baillot, A. & Busch, A., 2014. "Berliner Intellektuelle um 1800" als Programm. Über Potential und Grenzen digitalen Edierens. *Romantik Digital*, 1 9.

Baillot, A. & Busch, A., 2015. Editing for man and machine: The digital edition Letters and texts. Intellectual Berlin around 1800 as an example. *Variants,* Volume 13.

Baillot, A., ed., n.d. *Letters and texts. Intellectual Berlin around 1800.* s.l., Berlin: Humboldt-Universität zu Berlin.

Baillot, A. & Schnöpf, M., 2015. Von wissenschaftlichen Editionen als interoperable Projekte, oder: Was können eigentlich digitale Editionen?. *Historische Mitteilungen der Ranke-Gesellschat,* Volume Beiheft 91, pp. 139-156.

Beckett, S., 2011-2018. *Digital Manuscript Project. A digital genetic edition,* Brussels: University Press Antwerp.

Bishop, C. M., 2006. *Pattern Recognition and Machine Learning (Information Science and Statistics).* s.l.:Springer Science+Business Media, LLC.

Blei, D. M., Griffiths, T. L. & Jordan, M. I., 2010. The Nested Chinese Restaurant Process and Bayesian Nonparametric Inference of Topic Hierarchies. *J. ACM,* Volume 57, pp. 1-30.

Blei, D., Ng, A. & Jordan, M., 2003. Latent Dirichlet allocation. *JMLR.*

Bojanowski, P., Grave, E., Joulin, A. & Mikolov, T., 2017. Enriching Word Vectors with Subword Information. *Transactions of the Association for Computational Linguistics,* Volume 5, pp. 135-146.

Carpenter, B., 2010. Integrating out multinomial parameters in latent Dirichlet allocation and naive Bayes for collapsed Gibbs sampling. *Rapport Technique,* Volume 4, p. 464.

Dalen-Oskam, K., 2016. Corpus-based approaches to Names in Literature. In: C. Hoigh, ed. *The Oxford Handbook of Names and Naming.* Oxford: Oxford University Press.

Deerwester, S. et al., 1990. Indexing by latent semantic analysis. *Journal of the American society for information science,* Volume 41, p. 391.

Ehrmann, D., 2016. Textrevision --- Werkrevision. Produktion und Überarbeitung im Wechsel von Autoren, Herausgebern und Schreibern. *Editio,* Volume 30.

Goethe, J. W., 2017. *Faust. Historisch-kritische Edition..* Frankfurt am Main / Weimar / Würzburg: s.n.

Gruber, A., Rosen-Zvi, M. & Weiss, Y., 2007. *Hidden Topic Markov Models.* s.l., JMLR.

Jannidis, F. et al., 2015. *Automatische Erkennung von Figuren in deutschsprachigen Romanen.* Graz: ADHO.

Jelodar, H. et al., 2019. Latent Dirichlet allocation (LDA) and topic modeling: models, applications, a survey. *Multimedia Tools and Applications,* 78(11), pp. 15169-15211.

Leclerc, Y., ed., 2009. *Les Manuscrits de Madame Bovary.* Rouen, s.n.

Liu, L. et al., 2016. An overview of topic modeling and its current applications in bioinformatics. *SpringerPlus,* Volume 5.

Müller, K.-R.et al., 2001. An introduction to kernel-based learning algorithms. *IEEE Transactions on Neural Networks,* 12(2), pp. 181-201.

Nakajima, S., Kazuho, W. & Masashi, S., 2019. *Variational BAyesian Learning Theory.* s.l.:Cambridge University Press.

Paisley, J., Wang, C., Blei, D. M. & Jordan, M. I., 2015. Nested hierarchical Dirichlet processes. *IEEE transactions on pattern analysis and machine intelligence,* Volume 37.

Plachta, B., 2006. *Editionswissenschaft: eine Einführung in Methode und Praxis der Edition neuerer Texte.* 2nd ed. Stuttgart: Reclam.

Ralle, I. H., 2016. Maschinenlesbar --- menschenlesbar. Über die grundlegende Ausrichtung der Edition. *Editio,* Volume 30.

Ramsay, S., 2011. *Reading Machines: Toward and Algorithmic Criticism.* s.l.:University of Illinois Press.

Rasmussen, C. E. & Williams, C. K. I., 2006. *Gaussian Processes for Machine Learning.* s.l.:MIT Press.

Rosen-Zvi, M., Griffiths, T., Steyvers, M. & Smyth, P., 2004. *The author-topic model for authors and documents.* s.l., s.n., pp. 487-494.

Samek, W. et al. eds., 2019. *Explainable AI: Interpreting, Explaining and Visualizing Deep Learning.* s.l.:Lecture Notes in Artificial Intelligence. Volume 11700.

Schöch, C., 2017. Topic Modeling Genre: An Exploration of French Classical and Enlightenment Drama. *Digital Humanities Quarterly,* 11(2).

Schlitz, S., 2014. Digital Texts, Metadata, and the Multitude. New Directions in Participatory Editing. *Variants,* Volume 11, pp. 71–-89.

Schmidt, D., 2016. Using standoff properties for marking-up historical documents in the humanities. *Information Technology: Human Computation,* Volume 58, pp. 63-69.

Shillingsburg, P., 2013. Development Principles for Virtual Archives and Editions. *Variants,* pp. 61-76.

Siemens, R. et al., 2012. Toward modeling the social edition: An approach to understanding the electronic scholarly edition in the context of new and emerging social media*. *Literary and Linguistic Computing,* Volume 27, pp. 445-461.

Wainwright, M. J. & Jordan, M. I., 2008. *Graphical Models, Exponential Families, and Variational Inference.* s.l.:now publishers.

Wallach, H., 2006. *Topic Modeling: Beyond Bag-of-words.* New York, ACM.

Witkowski, G., 1924. *Textkritik und Editionstechnik neuerer Schriftwerke.* Leipzig: Haessel.

Xuan, J., Lu, J., Zhang, G. & Luo, X., 2015. Topic model for graph mining. *IEEE transactions on cybernetics,* Volume 45.

# APPENDIX

## NON-CONTENT-RELATED ALTERATION PROCESSING

To identify the non-content-related alteration categories, existing methods are used, however, for the identification of the specific content related reasons on the right, the novel alterLDA method is used. As the main contribution of this work lays in introducing the new alterLDA model, the description of the non-content-related alterations. In this section, the description of the established methods is shortly summarized, whereas the description of alterLDA is given more space in the forthcoming subsections.

### *PARATEXTS*

In Figure 2, the excerpt of the facsimile marked as archival note (orange) has the number 6 written in the top right corner of the sheet, this detail is shown in Figure 9. The corresponding xml transcription is the following:

```
<NOTE TYPE="FOLIATION" PLACE="MARGIN-RIGHT INLINE" HAND="#PENCIL_1">6</NOTE>
```

As the header reveals, this pencil numbering has been performed by an archivist:

```
<handNote xml:id="pencil_1" scope="minor" medium="pencil"
scribe="archivist">
    <seg xml:lang="de">Hand eines Archivars, in Bleistift.</seg>
    <seg xml:lang="en">Hand of an archivist, in pencil.</seg>
    <seg xml:lang="fr">Main d'un archiviste, crayon de papier.</seg>
</handNote>
```

For the second pencil note in Figure 9 there is no scribe annotated although it also contains nothing but a numbering:

```
<note type="foliation" place="align(center)" hand="#pencil_2">
    <hi rend="underline">99</hi>
</note>
```

The additions that have been performed by a different hand than the primary author and that contain numberings or dates, we consider to be archivists notes. Such archivist's or editor's additions can be identified with a very basic set of rules.

*Figure is not included in the preprint as the permission to publish the image sources only covers the journal article*

*Figure 9: Archival note. BI, Adelbert von Chamisso to Louis de La Foye. Nachlass 239, Blatt 6. Staatsbibliothek Berlin / Manuscripts section. Reuse subject to prior approval by Staatsbibliothek Berlin.*
*Published in: Letter from Adelbert von Chamisso to Louis de La Foye (fragment) (without place, 26 june 1804). Ed. by Anna Busch, Sabine Seifert. Prepared by Janine Katins. In collaboration with Sabine Seifert, Sophia Zeil. In: "Letters and texts: Intellectual Berlin around 1800". Ed. by Anne Baillot. Berlin: Humboldt-Universität zu Berlin. http://www.berliner-intellektuelle.eu/manuscript?Brief005ChamissoandeLaFoye. Last modified: 27 April 2015.*

### CORRECTIONS

*Figure is not included in the preprint as the permission to publish the image sources only covers the journal article*

*Figure 10: Correction of mistake of "wurde" to "würde". BI, Adelbert von Chamisso to Louis de La Foye. Nachlass 239, Blatt 85. Staatsbibliothek Berlin / Manuscripts section. Reuse subject to prior approval by Staatsbibliothek Berlin. Published in: Letter from Adelbert von Chamisso to Louis de La Foye (Geneva, at the beginning of 1812). Ed. by Anna Busch, Sabine Seifert. Prepared by Lena Ebert. In: "Letters and texts: Intellectual Berlin around 1800". Ed. by Anne Baillot. Berlin: Humboldt-Universität zu Berlin. http://www.berliner-intellektuelle.eu/manuscript?Brief047ChamissoandeLaFoye. Last modified: 27 April 2015.*

The alteration marked in green replaces a single character of a word, for which the corresponding part of the facsimile is shown in Figure 10.

*[..]Geschichte wuürde lang*

*und schal ausfallen[..]*

*BI, Adelbert von Chamisso to Louis de La Foye. Letter 47, p. 1*

This alteration is a correction of a mistake and conceptually, it is worth noting that the words before and after the alteration are very similar. This characteristic will be exploited for the identification of corrections. Identifying corrections is a considerably more difficult task because the corrected version does not necessarily have to be correct from what we know today. The fact that the alteration author corrected the text only means that he or she thought that his or her version is correct. We thus cannot rely on comparing the second version of the text with what an automatic spell checker would the first version correct to. Instead, we divided the problem even further into spelling alterations and grammatical alterations. For identifying spelling mistakes, the tokens of both versions are fuzzy-string matched against the common dictionary of lemmas. If both tokens match closely to the same lemma according to the Levenshtein distance, the two tokens are considered two different spellings of the same word. Fuzzy string matching of multiple tokens against a large vocabulary can be costly in terms of computing time and memory. For a larger data set an adjustment to this approach may be necessary. However, this approach gave better results than simply comparing Levenshtein distance of the tokens of both versions with each other, due to smaller tokens that are very similar but mean different things (e.g. “hate” and “fate” have Levenshtein distance of 1 but have a very different meaning.) For identifying grammatical alterations, we assume that the forms of the tokens in the sentence change and probably punctuations are added or deleted, but the set of lemmas is preserved for the most part. Hence, if the forms or the part of speech of the tokens in the span change but the set of lemmas do not, this alteration is a grammatical correction.

### *Stylistic Alterations*

*[..] DaherEs bedarf esdaher hier eines [..]*

*BI, About the notion of philosophy PP. by Immanuel Hermann von Fichte, p. 14*

For identifying stylistic alterations, we assume that all corrections and paratexts are already labelled according to the described method. Thus, there are only stylistic alterations and moral censorships left to be labelled. In our understanding, a stylistic alteration preserves the meaning of the text by only changing the way it is posed which includes rearranging of words, the use of synonyms and rephrasing. In recent years, a method gained a lot of attention that strive to find a vector-space representation of words that capture its meaning. Words or sentences projected to this space reveal a high similarity (for example cosine-similarity) if they have the same meaning. We introduce a threshold and consider all alterations for which the vector-space embedding of the text before and after the alteration reveal a smaller distance to be a stylistic alteration.

The alteration marked in blue (Figure 2) which is shown in higher resolution in Figure 11, reorders the words at the beginning of a sentence without changing the meaning.

*Figure is not included in the preprint as the permission to publish the image sources only covers the journal article*

*Figure 11: Stylistic alteration of "Daher bedarf es" to "Es bedarf daher". Immanuel Hermann Fichte: Über den Begriff der Philosophie, Humboldt-Universität zu Berlin, Universitätscharchiv, HU UA, Phil.Fak.01.Prom., No. 210, Page 14. Reuse subject to prior approval by the Universitätsarchiv.Ed. by Eva Schneider. Prepared by Eva Schneider, Anne Baillot, Denny Becker. In collaboration with Johanna Preusse. In: "Letters and texts: Intellectual Berlin around 1800". Ed. by Anne Baillot. Berlin: Humboldt-Universität zu Berlin. http://www.berliner-intellektuelle.eu/manuscript?IHFichte_Die_Aufgabe_der_Philosophie. Last modified: 12 January 2015.*

For completeness, we also provide the individual facsimile of the content related alteration example in Figure 12.

*Figure is not included in the preprint as the permission to publish the image sources only covers the journal article*

*Figure 12: Content-related alteration. Nachlass Uechtritz. Oberlausitzische Bibliothek der Wissenschaften Görlitz. Reuse subject to prior approval by Oberlausitzische Bibliothek der Wissenschaften Görlitz. Letter from Dorothea Tieck to Friedrich von Uechtritz (Dresden, 10 April 1835). Ed. by Sophia Zeil. Published in: "Letters and texts: Intellectual Berlin around 1800". Ed. by Anne Baillot. Berlin: Humboldt-Universität zu Berlin. http://www.berliner-intellektuelle.eu/manuscript?Brief16DorotheaTieckanUechtritz. Last modified: 24 January 2015.*

*Used Symbols*

| Identifier | Meaning | Type | Dimensionality |
|---|---|---|---|
| V | Number of unique tokens in the dictionary | Int | |
| W | Number of tokens in the corpus | Int | |
| M | Number of documents | Int | |
| $N_m$ | Number of tokens in document m | Int | |
| K | Number of topics | Int | |
| $\alpha$ | Concentration of $\theta$ | Hyper parameter | K |
| $\eta$ | Concentration of $\beta$ | Hyper parameter | V |
| $\xi$ | Concentration of $\gamma$ | Hyper parameter | 2 |
| $\beta$ | Topic-term variable | Dirichlet | K x V |
| $\theta$ | Document-topic variable | Dirichlet | M x K |
| $\gamma$ | Topic-alteration-tendency variable | Dirichlet | K x 2 |
| $z$ | Token-topic variable | Categorical | W x K |
| $w$ | Tokens | Observed (Categorical) | W x V |
| $c$ | Alteration | Observed (Categorical) | W X 2 |

## Collapsed Gibbs Sampler of the AlterLDA Model

For the Collapsed Gibbs Sampler of the alterLDA model it is shown how to derive the posterior for the topic assignment at a current position, given the current configuration. First, the joint probability of the whole model is given before showing how to compute the topic assignment based on count statistics. The joint probability of the model is given by

$$
\begin{aligned}
p(\mathbf{w}, c, z, \gamma, \beta, \theta \mid \alpha, \eta, \xi) = \; & p(c \mid z) \cdot p(\mathbf{w} \mid z, \beta) \cdot p(\gamma \mid \xi) \cdot p(\beta \mid \eta) \cdot p(z \mid \theta) \cdot p(\theta \mid \alpha) \\
= \; & \prod^{M,N} \mathrm{CAT}(c \mid z, \gamma) \times \prod^{M,N} \mathrm{CAT}(\mathbf{w} \mid z, \beta) \times \prod^{M,N} \mathrm{CAT}(z \mid \theta) \\
& \times \prod^{M} \mathrm{DIR}(\theta \mid \alpha) \times \prod^{K} \mathrm{DIR}(\gamma \mid \xi) \times \prod^{K} \mathrm{DIR}(\beta \mid \eta)
\end{aligned}
$$

We introduce a counter variable **c** which can be indexed in four dimensions, the current topic ($k$), the current document ($m$), the current alteration mode ($a$) and the current token ($\mathbf{w}$).

$$\mathbf{c}_{k,m,a,\mathbf{w}} = \sum_{n=1}^{N_m} \mathbf{I}\,(z_{m,n} = k \quad \& \quad w_{m,n} = \mathbf{w} \quad \& \quad c_{m,n} = a)$$

In this setting, the desired computation is the probability of a topic assignment at a specific position given a current configuration of all other topic assignments. This probability can be formalized by

$$p(z_{m,n} \mid z_{-(m,n)}, \mathbf{w}, c, \alpha, \eta, \xi) \propto p(z_{m,n}, z_{-(m,n)}, \mathbf{w}, c \mid \alpha, \eta, \xi)$$

Adopting Equation 16 from the Carpenter paper, this probability can be written by marginalizing $\theta$, $\beta$ and $\gamma$ from the joint probability.

$$\begin{aligned} p(z_{m,n}, z_{-(m,n)}, \mathbf{w}, c \mid \alpha, \eta, \xi) = \; & \int\int\int p(\mathbf{w}, c, z, \gamma, \beta, \theta \mid \alpha, \eta, \xi)\; d\theta d\beta d\gamma \\ = \; & \underbrace{\int p(\theta \mid \alpha) \cdot p(z \mid \theta)\; d\theta}_{A} \\ & \times \underbrace{\int p(\mathbf{w} \mid z, \beta) \cdot p(\beta \mid \eta)\; d\beta}_{B} \\ & \times \int p(c \mid z, \gamma) \cdot p(\gamma \mid \xi)\; d\gamma \\ = \; & \prod_{k=1}^{K} \int p(\gamma_k \mid \xi) \prod_{m=1,n=1}^{M,N_m} p(c \mid \gamma_{\mathrm{argmax}(z_{m,n})})\; d\gamma_k \times A \times B \end{aligned}$$

A and B are substituted here because their derivation is identical to the one in Carpenter et al. Analogue to Equation 27 of Carpenter et al., after inserting the definitions of the Dirichlet distribution the result is proportional to three factors.

$$\propto \quad \left(\mathbf{c}^{-}_{z_{m,n,*,*}} + \alpha_{z_{m,n}}\right)\left(\frac{\mathbf{c}^{-}_{z_{m,n},*,*,\mathbf{w}_{m,n}} + \eta_{w_{m,n}}}{\mathbf{c}^{-}_{z_{m,n},*,*,*} + \sum_v^V \eta_v}\right)\left(\frac{\mathbf{c}^{-}_{z_{m,n},*,c_{m,n},*} + \xi_{w_{m,n}}}{\mathbf{c}^{-}_{z_{m,n},*,*,*} + \sum_i^2 \xi_i}\right)$$

Where $\cdot^{-}$ denotes the counter disregarding the current position $m, n$.

## Results on Synthetic Data

A big advantage of generative models is that they can be used to generate new data. The generated documents themselves may not be too interesting in the case of topic models, but they can be used to evaluate the functionality of the model. To do this, first, the variables of the model are initialized, and documents are generated. Now the variables are initialized again and based on the previously generated documents the old variable configurations are reconstructed. The performance of the inference can be measured by the accuracy of the reconstruction.

In Figure 13, the results of such an evaluation over 324 different experiment runs is shown, within the sparsity of the hyper-parameters as well as the number of tokens were varied. Each cell shows the mean reconstruction of c over two runs with a given set of parameter choices. The brighter the color of the cell, the better the reconstruction. Overall, a smaller alpha yields better results, independent of the size of the data set and the choice of the other concentration factors. Intrestingly, for fewer documents and $\alpha = 0.1$, a smaller concentration factor $\eta$ performs better, wheras for either a larger number of documents or a larger $\alpha = 1.0$, a larger $\eta$ is to be preferred. The explanation for this result is that with more documents, the exact proportions of the topics can be inferred more accurately, wheras for fewer documents, there is the chance of getting a few (sparse) topics right.

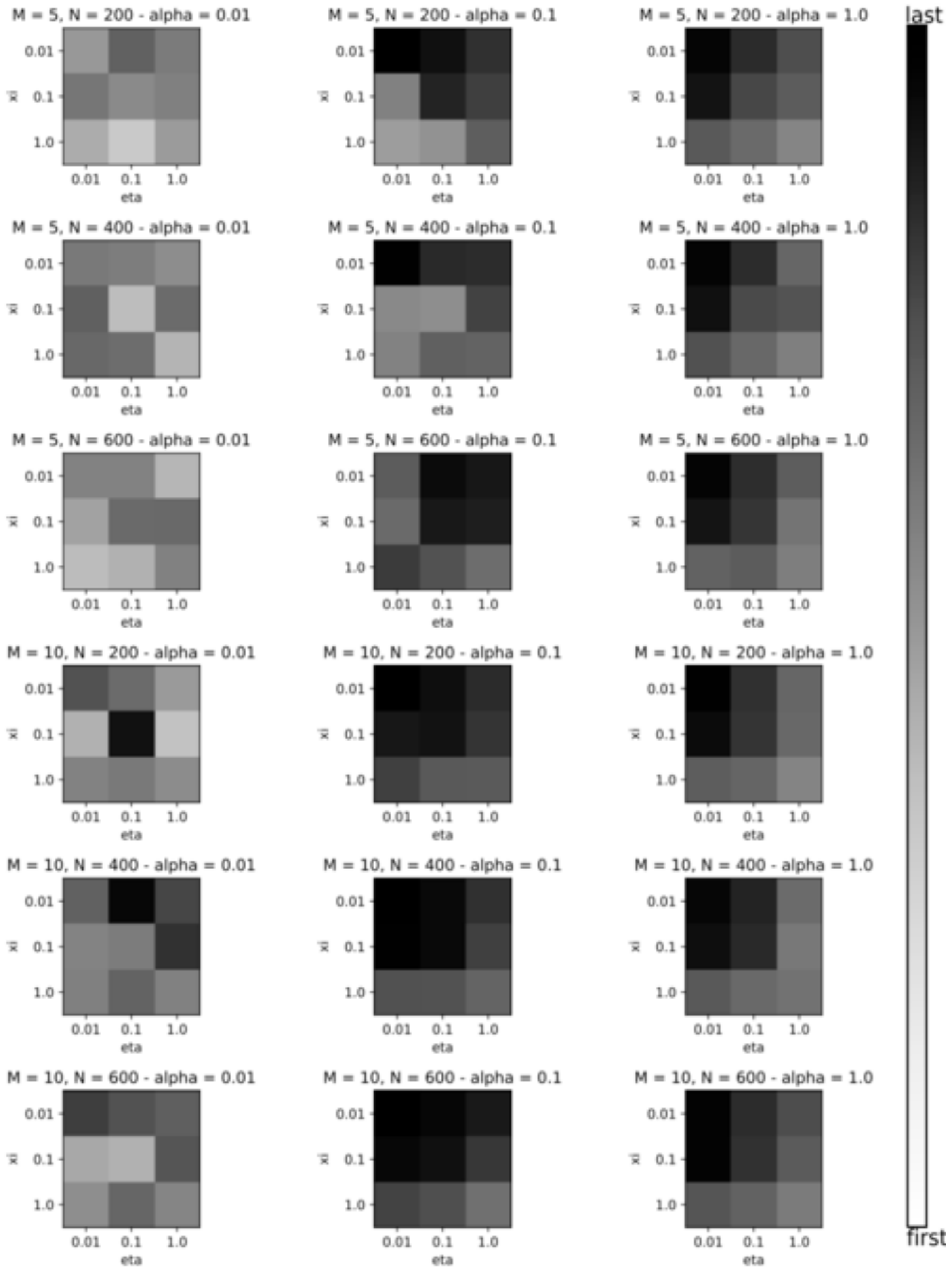

*Figure 13: Grid search result for training accuracy of $\hat{c}$ parameter on synthetic data. Even with a small total number of tokens, the accuracy can be very high. Interestingly, the accuracy depends strongly on the sparsity of $\alpha, \xi$ and $\eta$.*

[a] David Lassner
Machine Learning Group, Technische Universität Berlin, 10587 Berlin, Germany

[b] Anne Baillot
Le Mans Université. 72085 Le Mans, France

[c] *Sergej Dogadov*
Machine Learning Group, Technische Universität Berlin, 10587 Berlin, Germany

[d] *Klaus-Robert Müller*
Machine Learning Group, Technische Universität Berlin, 10587 Berlin, Germany
Berlin Big Data Center, 10587 Berlin, Germany
Department of Brain and Cognitive Engineering, Korea University, Anam-dong, Seongbuk-gu, Seoul 136-713, South Korea
Max-Planck-Institut für Informatik, Saarbrücken, Germany
Berliner Zentrum für Maschinelles Lernen, 10587 Berlin, Germany

[e] *Shinichi Nakajima*
Machine Learning Group, Technische Universität Berlin, 10587 Berlin, Germany
Berlin Big Data Center, 10587 Berlin, Germany
RIKEN Center for AIP, 103-0027, Tokyo, Japan

# NOTES

[f] The encoding guidelines can be found at berliner-intellektuelle.eu/encoding-guidelines.pdf
[g] gitlab.tubit.tu-berlin.de/david.lassner/shipping_alterLDA
[h] „[..] Tieck plante aus finanzieller Bedrängnis heraus den Verkauf seiner Bibliothek an den Grafen Yorck von Wartenburg[..]" BI, comment by Johanna Preusse in letter from Ludwig Tieck to Friedrich von Raumer (Dresden, 11. November 1836)
[i] „Kontext der von Tieck angedeuteten Vorgänge waren Machtstreitigkeiten zwischen der katholischen Kirche und dem preußischen Staat. [..]" BI, comment by Johanna Preusse in letter from Ludwig Tieck to Friedrich von Raumer (Dresden, 27. März 1838)